\pdfoutput=1

\documentclass[11pt]{article}

\usepackage[final]{acl}

\usepackage{times}
\usepackage{latexsym}

\usepackage{microtype}
\usepackage{hyperref}
\usepackage{url}
\usepackage{booktabs}
\usepackage{wrapfig}
\usepackage{graphicx}
\usepackage{multirow}
\usepackage{tcolorbox}
\usepackage{tabularx}
\usepackage{amsmath}
\usepackage{amssymb}

\usepackage{pifont}

\definecolor{darkgreen}{rgb}{0.0,0.5,0.0}
\newcommand{\cmark}{\textcolor{darkgreen}{\ding{51}}}
\newcommand{\xmark}{\textcolor{red}{\ding{55}}}

\usepackage[T1]{fontenc}

\usepackage[utf8]{inputenc}

\usepackage{microtype}

\usepackage{inconsolata}

\usepackage{graphicx}

\title{Re-ReST: Reflection-Reinforced Self-Training for Language Agents}

\author{Zi-Yi Dou, Cheng-Fu Yang , Xueqing Wu , Kai-Wei Chang , Nanyun Peng \\ \textit{University of California, Los Angeles} \\ \texttt{\{zdou,cfyang,xueqing.wu,kwchang,violetpeng\}@cs.ucla.edu} 
}

\begin{document}
\maketitle
\begin{abstract}
Finetuning language agents with reasoning-action trajectories is effective, but obtaining these trajectories from human annotations or stronger models is costly and sometimes impractical. In this paper, we investigate the use of self-training in language agents, which can generate supervision from the agent itself, offering a promising alternative without relying on human or stronger model demonstrations. Self-training, however, requires high-quality model-generated samples, which are hard to obtain for challenging language agent tasks. To address this, we present Reflection-Reinforced Self-Training (Re-ReST), which uses a \textit{reflector} to refine low-quality generated samples during self-training. The reflector takes the agent's output and feedback from an external environment (e.g., unit test results in code generation) to produce improved samples. This technique enhances the quality of inferior samples and efficiently enriches the self-training dataset with higher-quality samples. We conduct extensive experiments on open-source language agents across tasks, including multi-hop question answering, sequential decision-making, code generation, visual question answering, and text-to-image generation. The results demonstrate the effectiveness of self-training and Re-ReST in language agent tasks, with self-training improving baselines by 7.6\% on HotpotQA and 28.4\% on AlfWorld, and Re-ReST further boosting performance by 2.0\% and 14.1\%, respectively. Our studies also confirm the efficiency of using a reflector to generate high-quality samples for self-training. Moreover, we demonstrate a method to employ reflection during inference without ground-truth feedback, addressing the limitation of previous reflection work. Our code is released at \url{https://github.com/PlusLabNLP/Re-ReST}.

\end{abstract}

\section{Introduction}

\begin{figure}
    \centering
    \includegraphics[width=0.45\textwidth]{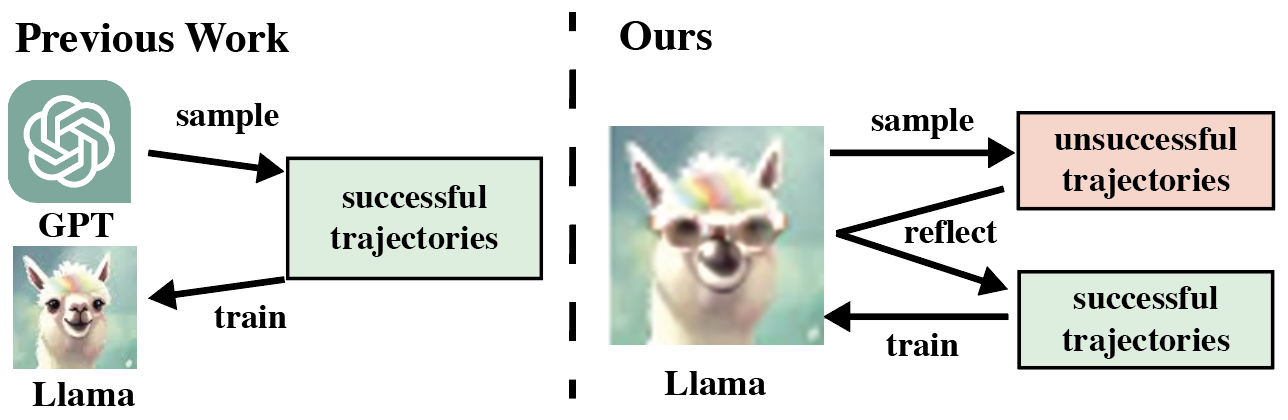}
\caption{\label{fig:fig1} Previous agent training methods~\cite{chen2023fireact,yin2024lumos} distill knowledge from stronger models (e.g., GPT-4) to weaker ones (e.g., Llama-2). In contrast, we adopt self-training and improve it with reflection to improve agents more autonomously, which reduces reliance on external propriety models and maintains a fully open-source framework.}
\end{figure}

Large language models (LLMs)~\citep{kenton2019bert,touvron2023llama,achiam2023gpt} have demonstrated potential in interacting with external environments and addressing practical interactive tasks, resulting in a new class --- language agents~\citep{nakano2021webgpt,yao2022react}. Finetuning LLMs for agentic tasks has proven effective, yet existing works rely on data generated by stronger models (e.g., GPT-4)~\citep{chen2023fireact,yin2024lumos}, which are not always available (e.g., to improve the strongest model). 

Among the potential techniques to improve agents~\citep{ouyang2022training,wang2023self,li2024self,chen2024self}, self-training holds promise for enhancing agent performance for challenging agentic tasks. The self-training process typically involves refining the model by generating samples, assessing their quality through rewards, and updating the model by training on high-quality samples. Compared with existing agent training methods~\cite{chen2023fireact,yin2024lumos}, self-training can autonomously improve agents and reduce the discrepancy between the agent's training data and its original predictions. Additionally, as in Figure~\ref{fig:fig1}, self-training can potentially allow for the development of performant agents within a fully open-source framework, without relying on closed-source, proprietary models. Given these benefits, we propose to investigate the use of self-training in language agents in this paper. 

\begin{figure*}
    \centering
    \includegraphics[width=1.0\textwidth]{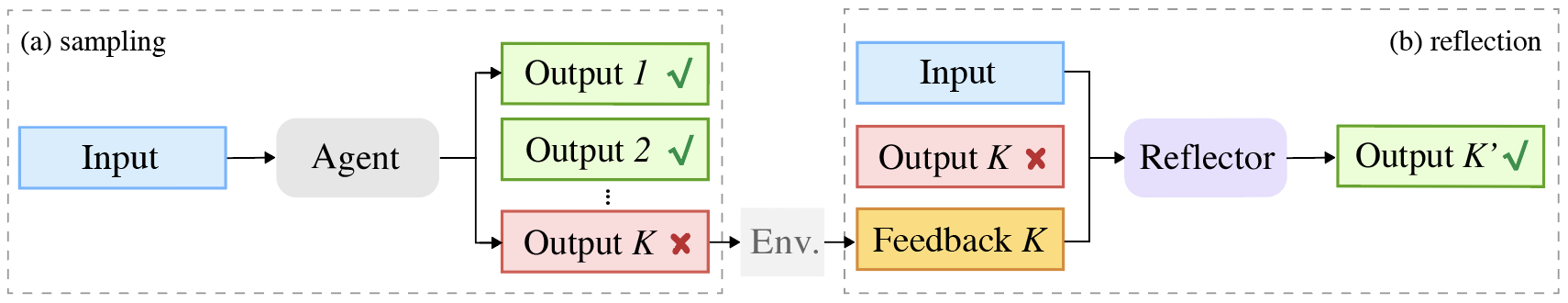}
\caption{\label{fig:overview} An overview of our Re-ReST method. Our approach incorporates self-training in language agent tasks by sampling multiple outputs from an agent and using positive samples for training. To enhance the effectiveness of self-training in language agents, we introduce a reflector mechanism. If a sample is incorrect, the reflector adjusts the agent's output based on environmental feedback. The corrected sample is then incorporated into the training data, thereby improving the overall self-training process. }
\end{figure*}

However, one significant challenge for applying self-training in language agent tasks lies in the acquisition of high-quality samples to achieve good performance. Specifically, self-training requires a substantial amount of high-quality samples, while relying solely on model-generated samples can be inefficient, particularly for language agent tasks that demand multi-step reasoning and long-horizon planning. As a result, it is challenging to obtain good samples solely through sampling. 
Moreover, the common practice of discarding low-quality samples neglects their potential for improvement and effective utilization, thus limiting the overall efficacy of self-training methods.

To address these issues, we propose Reflection-Reinforced Self-Training (Re-ReST), which enhances the self-training algorithm using a reflection model. Re-ReST incorporates a \textit{reflector} during self-training, which improves sample quality by utilizing environmental feedback such as execution successes and unit test outcomes. Specifically, the reflector transforms lower-quality samples into higher-quality ones, leveraging the capability of LLMs to self-improve when provided with accurate ground-truth feedback~\cite{huang2024large}. Consequently, it enriches the training dataset, enabling more effective bootstrapping. After training, only the agent model is used for inference, ensuring no additional computational burden during testing. Unlike existing self-reflection methods~\citep{madaan2023self,shinn2023reflexion,pan2023automatically}, Re-ReST only requires access to feedback during training, not during inference, making our setting more realistic and practical.

We conduct extensive experiments with open-source LLMs across a wide range of tasks, including multi-hop question answering, sequential decision-making, code generation, visual question answering, and text-to-image generation. Our results first demonstrate the potential of self-training in language agent tasks, showing improvements over few-shot baselines in long-horizon planning tasks, with gains of 7.6\% on HotpotQA and 28.4\% on AlfWorld. By incorporating Re-ReST, we further enhance performance significantly by 2.0\% and 14.1\% on HotpotQA and AlfWorld, respectively, achieving results better or comparable to models relying on commercial APIs. Ablation studies confirm the efficiency of the reflection model in generating high-quality self-training samples. Furthermore, we explore using our reflection model during inference with self-consistency decoding, which improves the model performance while alleviating the need for ground-truth feedback required by previous work~\cite{huang2024large}. Additionally, we demonstrate the application of our method in preference optimization objectives.

\section{Method: Re-ReST}

\paragraph{Self-Training.} Formally, given a dataset $U=\{x_i\}_{i=1}^{N}$, self-training begins by using a base model $\mathcal{M}$ to generate a pseudo-label $\hat{y}_i = \mathcal{M}(x_i)$ for each instance $x_i \in U$. Subsequently, a subset of $\{ (x_i, \hat{y}_i ) \}_{i=1}^N$ is selected based on a scoring function, and $\mathcal{M}$ is finetuned on this selected subset. For language agents, we define the label $y$ as a trajectory comprising interleaved thoughts and actions, as described in ReAct~\cite{yao2022react}. We propose adopting the self-training paradigm by training language agents with their self-generated thought-action trajectories.

\paragraph{Overview of Re-ReST.} Obtaining high-quality samples through self-sampling can be challenging, particularly for complex language agent tasks. To address this issue, we introduce Re-ReST, which aims to enhance the pseudo-label generation process in self-training for language agents. As illustrated in Figure~\ref{fig:overview}, we propose improving low-quality samples using a reflection model with external feedback. We then enrich the self-training data by incorporating these corrected generations. This process generates high-quality samples efficiently by correcting low-quality ones with ground-truth feedback during training.

\subsection{Components}

Our method involves two models, including a language agent $\mathcal{M}$ that generates text and actions, and a reflection model $\mathcal{R}$ that improves a low-quality sample. The reflection model $\mathcal{R}$ has access to an external environment $\mathcal{E}$ that can provide external feedback to a generated sample (e.g. numerical scores and/or verbal error information). We illustrate each of these modules in the following part.

\paragraph{Language Agent.} The language agent $\mathcal{M}$ is built upon a large language model (LLM) that is trained or prompted to generate thoughts and actions given a task. Formally, given an instance $x_i$, the agent $\mathcal{M}$ generates its output $\hat{y} \sim \mathcal{M}(\mathbf{y}|x)$ containing its actions. The agent can first generate its reasoning traces before outputting its actions, which has been demonstrated to improve the model performance and interpretability~\citep{yao2022react}.

\paragraph{Reflector.} The reflection model $\mathcal{R}$ is also instantiated as an LLM, the goal of which is to improve the language agent's generations given external feedback. We assume that during training, an external environment $\mathcal{E}$ can evaluate a generated sample and provide feedback $\mathcal{E}(x, \hat{y})$ to the agent. The feedback can be a binary success status and/or error information. For example, in code generation tasks, the environment can execute the model-generated code on unit tests, providing information on whether the code has syntax errors and whether it can pass the unit tests. Having access to such an environment is important in our setting, as it has been shown that an LLM cannot perform self-correction without high-quality external feedback~\citep{huang2024large}. The reflection model generates a corrected sample $\tilde{y} \sim \mathcal{R}(\mathbf{y}|x, \hat{y}, \mathcal{E}(x, \hat{y}))$ given the task information $x$, the agent generation $\hat{y}$, and the environmental feedback $\mathcal{E}(x, \hat{y})$. It can optionally first state its reasoning process (e.g., which specific actions could be corrected) before generating the corrected answer.) The use of the reflection model can improve self-training by finding good solutions efficiently because of the additional information provided (i.e., the agent's previous trial and the environmental feedback.) We do not share the model parameters between the agent and reflector in this paper.

\subsection{Data Generation}

We then describe how we generate self-training data for the language agent $\mathcal{M}$. The data generation process involves two steps, including the initial generation step with the language agent itself and the reflection step with the reflector, and we obtain the agent-generated dataset $\mathcal{D}_{\mathcal{M}}$ and reflector-generated dataset $\mathcal{D}_{\mathcal{R}}$ from the two steps.

\paragraph{Initial Generation.} As in the standard setup, given an instance $x$, we sample $k$ generations $\{\hat{y}^j\}_{j=1}^k$ from the current language agent model $\hat{y}^j \sim \mathcal{M}(\mathbf{y}|x)$. Then, the environment $\mathcal{E}$ scores the generation and provides feedback $\mathcal{E}(x, \hat{y}^j))$. If the score exceeds a threshold, we add the instance to $(x, \hat{y}^j)$ to the training data $\mathcal{D}_{\mathcal{M}}$. In practice, we observe that setting $k=3$ achieves a good balance between efficiency and effectiveness.

\paragraph{Reflection with Environmental Feedback.} The initial generation step only relies on the agent model $\mathcal{M}$ itself to generate data. For a sampled generation $\hat{y}^j$, if the score does not pass the threshold, we will feed it to the reflection model for refinement. The reflector takes as inputs the task information $x$, the agent's prior generation $\hat{y}^j$, and the environmental feedback $\mathcal{E}(x, \hat{y}^j))$, and then generates the corrected sample $\tilde{y}^{j} \sim \mathcal{R}(x, \hat{y}^j, \mathcal{E}(x, \hat{y}^j)).$ The corrected sample $\tilde{y}^{j}$ will also be evaluated by the environment and we will add it to the reflector-generated training dataset $\mathcal{D}_{\mathcal{R}}$ if its score exceeds the threshold. While the reflection procedure can be iteratively applied multiple times as per~\citet{shinn2023reflexion}, in this study, we limit this process to a single iteration for the sake of efficiency. This means that each generated sample $\hat{y}^j$ is allowed a maximum of one refined counterpart $\tilde{y}^{j}$.

\subsection{Model Training and Inference}
\label{sec:train}

We first train the reflector $\mathcal{R}$ parameterized by $\theta_{\mathcal{R}}$ and then use the trained reflector to generate the reflection data $\mathcal{D}_{\mathcal{R}}$. Afterward, we combine $\mathcal{D}_{\mathcal{R}}$ and the agent's self-generated data $\mathcal{D}_{\mathcal{M}}$ to train the agent model $\mathcal{M}$ parameterized by $\theta_{\mathcal{M}}$.

\paragraph{Reflector Training.} While base LLMs can perform self-reflection or self-correction without any finetuning given ground-truth feedback~\citep{shinn2023reflexion}, we propose to further improve its reflection ability with the self-generated data. First, from the initial generation step, we obtain multiple generations $\{{y}^j\}_{j=1}^k$ from the agent model $\mathcal{M}$. For each correct generation ${y}^w$ and incorrect generation ${y}^l$ with its environmental feedback $\mathcal{E}(x, \hat{y}^l)$ in $\{{y}^j\}_{j=1}^k$, we will add the instance $\langle x, {y}^l, \mathcal{E}(x, \hat{y}^l), {y}^w \rangle$ to the agent-generated dataset $\mathcal{D}_{\mathcal{M}}^{\mathcal{R}}$ for reflector training. In addition, the reflector generates its self-training dataset in a zero-shot manner $\mathcal{D}_{\mathcal{R}}^{\mathcal{R}}$ similar to the agent initial generation step. Combining the two generated datasets, we train the reflector on $\mathcal{D}_{\mathcal{M}}^{\mathcal{R}} \cup \mathcal{D}_{\mathcal{R}}^{\mathcal{R}}$ with the standard maximum log-likelihood objective first before generating the training data $\mathcal{D}_{\mathcal{R}}$ for the language agent:
\begin{equation}
\small
  \mathcal{L}_{MLE}(\theta_\mathcal{R}) = - \mathbb{E}_{ (x, {y}^l, {y}^w) \sim \mathcal{D}_{\mathcal{M}}^{\mathcal{R}} \cup \mathcal{D}_{\mathcal{R}}^{\mathcal{R}}} \log p_{\theta_\mathcal{R}}({y}^w| x, {y}^l).
\end{equation}

\paragraph{Language Agent Training.} After we have the base language agent to generate the self-training data $\mathcal{D}_{\mathcal{M}}$ and the improved reflector to generate the reflector-generated data $\mathcal{D}_{\mathcal{R}}$, we train the language agent jointly on $\mathcal{D}_{\mathcal{M}} \cup  \mathcal{D}_{\mathcal{R}}$:

\begin{equation}
    \mathcal{L}_{MLE}(\theta_\mathcal{M}) = - \mathbb{E}_{ (x, {y}) \sim \mathcal{D}_{\mathcal{M}} \cup \mathcal{D}_{\mathcal{R}} }\log p_{\theta_\mathcal{M}}({y}| x).
\end{equation}

Besides the maximum log-likelihood objective, because the reflection training and data generation process involves the use of preference pairs, it is natural to use preference optimization objectives such as DPO~\citep{rafailov2024direct} for training, which we will discuss in the experiment section.

\paragraph{Inference.} During inference, accessing high-quality environmental feedback is often challenging, which can cause inference-time self-reflection algorithms to fail~\citep{huang2024large}. Therefore, we only have the agent $\mathcal{M}$ directly output generations without the reflector during inference. This approach eliminates the need for feedback and avoids any additional computational overhead. A potential method to integrate the reflector into the inference process involves first training a scorer to evaluate the agent's output. If the score falls below a certain threshold, self-correction can then be performed, which we leave as a future direction. Additionally, we propose performing reflection regardless of environmental feedback and employing self-consistency to derive the final results from both the agent's outputs and the reflector's outputs, as shown in the experiment section.

\section{Experiments}

We experiment with multi-hop reasoning, sequential decision-making, code generation, visual question answering, and text-to-image generation. 
We present the experimental settings and results for each task.
In all our experiments, we advocate for the use of open-source models and aim to avoid black-box, closed-source commercial models whenever possible.

\subsection{Multi-Hop Reasoning}
\paragraph{Dataset.} We use the HotpotQA dataset~\citep{yang2018hotpotqa}, a well-established question-answering dataset featuring multi-hop reasoning and knowledge retrieval. It is constructed based on Wikipedia and an agent needs to retrieve and reason over multiple supporting documents to answer a question. We sample 5,000 training instances randomly for self-training and 500 instances from the development set for evaluation as in~\citet{chen2023fireact}.

\paragraph{Model Setup.} We build both the agent model and the reflector upon the Llama-2-13B and Llama-3-8B models~\citep{touvron2023llama}. Note that different from previous work~\citep{shinn2023reflexion,chen2023fireact,yin2024lumos}, we do not employ a stronger language model such as GPT-3.5/4 for data generation or self-reflection, ensuring that the models do not benefit from knowledge distillation. Following~\citet{shinn2023reflexion}, we use the ReAct~\cite{yao2022react} method where at each step, the agent model first generates its thoughts and then performs an action. The action is chosen from (1) Search[entity], which searches the exact entity on Wikipedia, (2) Lookup[keyword], which localizes a keyword in the retrieved passages, and (3) Finish[answer], which returns the answer and finishes the task. We use a free Wikipedia API\footnote{\url{https://python.langchain.com/docs/integrations/tools/wikipedia}} for passage retrieval and keyword lookup. 

\paragraph{Training and Evaluation Setup.} We use 2-shot prompting for few-shot agent and reflector data generation as in~\citet{shinn2023reflexion}. For each training instance, the agent model samples 3 generations. The generation is evaluated with the exact match metric (i.e., if the generated answer is exactly the same as the ground-truth answer). The retrieval and evaluation results are given to the reflector as the environmental feedback for self-correction. We use Low-Rank Adaptation (LoRA)~\citep{hu2022lora} for training the language models for efficiency. The agent and reflector models are trained for 3 epochs with a learning rate of 3e-4.
\begin{table*}[t]
\begin{center}
\setlength{\tabcolsep}{8.0pt}
\begin{tabular}{lccc}
\toprule
\multicolumn{1}{c}{\multirow{2}{*}{\bf Model}}  &  \multirow{2}{*}{\bf QA Tool} & \multicolumn{1}{c}{\bf  \#Train Data} &  \multicolumn{1}{c}{\multirow{2}{*}{\bf EM}} \\
& &  \bf (Self/GPT-4 Generated) & \\
\midrule
\multicolumn{2}{l}{{ \it{Llama-2-13B ReAct-Based Agents}}} \\
Few-Shot & WikipediaAPI  & - & 20.0 \\
Self-Training  & WikipediaAPI   & 2k/0  & 27.6 \\
Re-ReST  & WikipediaAPI  & 2.5k/0  & 29.6\\
\midrule
\multicolumn{2}{l}{{ \it{Llama-2-13B ReAct-Based Agents w/ GPT-4-Generated Data}}} \\
{FireAct~\citep{chen2023fireact}} & SerpAPI & 0/0.5k & {34.4} \\
{LUMOS~\citep{yin2024lumos}} & GPT-3.5 & 0/20k & {31.4} \\
{LUMOS~\citep{yin2024lumos}} & GPT-4 & 0/20k & {36.3} \\
FireAct & WikipediaAPI & 0/0.5k &  32.2 \\
Self-Training  &WikipediaAPI  & 2.5k/0.5k & 34.2 \\
Re-ReST   &WikipediaAPI  & 3k/0.5k & 35.8 \\
\midrule
\multicolumn{2}{l}{{ \it{Llama-3-8B ReAct-Based Agents}}} \\
Few-Shot & WikipediaAPI  & - & 30.0 \\
Self-Training & WikipediaAPI    & 2.4k/0 & 34.4 \\
Re-ReST  &WikipediaAPI  &  3k/0 & 36.8 \\
\bottomrule
\end{tabular}
\end{center}
\caption{\label{results:qa} On HotpotQA, our method enables a better usage of the training data compared with self-training and improves self-training for LLama-2/3-based agents. Also, adding only ~0.5k GPT-generated data enables our agents with the free Wikipedia API to achieve comparable or better performance than methods with commercial APIs.}
\end{table*}

\paragraph{Main Results.} We list the main results in Table~\ref{results:qa}. As shown in the table, self-training can significantly improve the model performance from an EM score of 20.0\% to 27.6\% for Llama-2 and from 30.0\% to 34.4\% for Llama-3. However, only 37.1\% and 48.3\% of the training instances are correctly solved by the agent model and are used for self-training respectively. By integrating our reflector model into the process, the agent can solve more training instances and thus have more data for training the agent model, increasing the EM scores significantly. In addition to our implemented models, following previous work (FireAct~\citep{chen2023fireact} and LUMOS~\cite{yin2024lumos}) that use GPT-3.5/4 for data generation and model finetuning, we employ GPT-4 to generate ~0.5k instances and first train the agents with the GPT-4 generated data before self-training. Results demonstrate that 1) self-training is a stronger baseline than FireAct under a fair setting where the same QA tool is used; 2) we can achieve comparable or better performance of our model than these methods, even though both of them use strong knowledge retrieval models (i.e., SerpAPI\footnote{https://serpapi.com/} for FireAct and GPT-4 for LUMOS), which are costly and non-scalable. By contrast, we use the free Wikipedia API.

\subsection{Sequential Decision-Making}
\paragraph{Dataset.}
We also assess the proposed approach on sequential decision-making using ALFWorld~\citep{shridhar2020alfworld}. ALFWorld comprises a collection of text-based settings designed to test an agent's ability to complete multi-step tasks across diverse interactive environments. Following~\citet{yao2022react,shinn2023reflexion}, we operate under the assumption that the agents are devoid of any access to successful trajectories, relying solely on a binary indicator of task success or failure. Our evaluation encompasses testing the agent across 134 previously~\textit{unseen} environments, spanning six diverse tasks. These tasks range from locating concealed items and transporting objects to interacting with objects using other items.

\begin{table}[tp]
\begin{tabularx}{0.5\textwidth}{
    l
    >{\centering\arraybackslash}X
    >{\centering\arraybackslash}X
}
\toprule
\multicolumn{1}{c}{\bf Model}  &  \multicolumn{1}{c}{\bf Sample Acc.} &  \multicolumn{1}{c}{\bf Success Rate} \\
\midrule
Few-Shot & - & 8.9 \\
Self-Training  &  11.2  & 37.3 \\
Re-ReST & 48.0 & 51.4 \\
\bottomrule
\end{tabularx}
\caption{\label{results:alfworld} Results on the ALFWorld dataset. Re-ReST substantially increases the sampling accuracy and outperforms self-training in terms of success rate even upon employing a reflector.}
\end{table}

\paragraph{Model Setup.} We build the agent and the reflector upon the Llama2-7b~\citep{touvron2023llama}. At each step, the agent can either contemplate its next move or generate admissible actions for execution as in~\citet{yao2022react}. Following the heuristics outlined by~\citet{shinn2023reflexion}, we trigger the reflector model for self-reflection if the agent repeats an action with the same response over three cycles, or if it performs over 30 actions in an environment.

\paragraph{Training and Evaluation Setup.}
We use one-shot prompting instead of the two-shot prompting in~\citet{shinn2023reflexion} for the models so that we can better fit a trajectory into the context window of Llama-2. We train the agent and reflector models on the collected trajectories for 2 epochs with a learning rate of 2e-5 using LoRA. 

\paragraph{Results.} As shown in Table~\ref{results:alfworld}, it is evident that the base Llama model faces challenges in adapting to the experimental environment, but self-training can significantly improve the model performance. A significant point to highlight is that the model operates without access to complete trajectories during the experiment. Despite this limitation, it demonstrates a notable improvement in performance within unseen environments—increasing the success rate from 8.9\% to 37.3\% through the utilization of self-augmented trajectories. Furthermore, the implementation of the reflector contributes a 14.1\% uplift in success rates, which affirms the efficacy of our proposed method.

\subsection{Programming: Code Generation and Visual Question Answering}
\paragraph{Dataset.} For code generation, we experiment with the Python code writing task on MBPP~\citep{austin2021program} and visual programming on GQA~\citep{gqa}. The MBPP benchmark consists of around 1,000 Python programming problems, with each problem paired with unit test cases. We follow its official split for the training and test data. The availability of the training set and its provided unit test cases make it suitable for our reflector to reflect and correct the model-generated code. For GQA, we randomly sample a subset of 5,000 data points for training and 1,000 data for testing.

\paragraph{Model Setup.} We build both the agent model and the reflector upon the CodeLlama-13B model~\citep{roziere2023code}. For MBPP, following~\citet{roziere2023code}, the agent model is given the unit test cases during code generation. Similarly, the reflection model is given the agent generation and its unit test results as the environmental feedback, and then generates a corrected version. For GQA, following~\citet{vipergpt}, we build the agent by providing a pre-defined set of visual APIs (e.g. object detection) and prompt the model to generate code using the APIs.

\paragraph{Training and Evaluation Setup.} For MBPP, we use zero-shot and three-shot prompting for zero-shot agent and reflector data generation. For GQA, we follow the prompt in~\citet{vipergpt} for the model for sample generation. For both datasets, the agent model samples 3 generations per training instance as before. We do not use the provided ground truths for MBPP training for consistency with the other experimental settings. The agent and reflector models are trained for 3 epochs with a learning rate of 3e-4 using LoRA.

\begin{table}[t]
\begin{center}
\setlength{\tabcolsep}{0.0pt}
\begin{tabular}{lcccccc}
\toprule
\multicolumn{1}{c}{\multirow{2}{*}{\bf Model}}  &  \multicolumn{2}{c}{\bf  MBPP } &  \multicolumn{2}{c}{\bf  GQA} \\
&\bf Sample Acc.  & \bf P@1   & \bf Sample Acc.   & \bf Score   \\
\midrule
Zero-Shot & - & 48.6  & - & 40.9\\
Self-Training  &  66.9  & 54.5 & 44.7 & 41.9\\
Re-ReST  & 77.3 & 56.4 & 55.7 & 42.6\\
\bottomrule
\end{tabular}
\end{center}
\caption{\label{results:code} Re-ReST improves self-training on code generation and visual programming tasks.}
\end{table}

\paragraph{Results.} As in Table~\ref{results:code}, for MBPP, because CodeLlama is trained on a large amount of code generation corpus, the base CodeLlama model can achieve a decent performance without any finetuning. The high pass rate results in many of the training instances being used for self-training. After self-training on the MBPP training data, the model performance can be improved from 48.6\% to 54.5\%. The reflector model can generate more self-training data and the pass rate can be improved with the reflector-generated data. For GQA, similar improvements can be seen, indicating that our method is also applicable in visual programming.

\subsection{Text-to-Image Generation}

\begin{table*}[t]
\begin{tabularx}{1.0\textwidth}{
    l
    >{\centering\arraybackslash}X
    >{\centering\arraybackslash}X
    >{\centering\arraybackslash}X
    >{\centering\arraybackslash}X
}
\toprule
\multicolumn{1}{c}{\multirow{2}{*}{\bf Model}} &  \multicolumn{1}{c}{\multirow{2}{*}{\bf Sample Acc.  }} &  \multicolumn{3}{c}{\bf VPEval Skill Score  } \\
& & \bf Count & \bf Spatial & \bf Scale  \\
\midrule
VPGen & - &  72.2 & 56.1 & 26.3  \\
VPGen w/ Self-Training &  57.6 &  74.7  &  54.5 & 29.3   \\
VPGen w/ Re-ReST  & 67.6 & 75.0 & 58.2 & 30.1  \\
\bottomrule
\end{tabularx}
\caption{\label{results:t2i} Re-ReST can outperform self-training in text-to-image generation when applied to VPGen and evaluated with VPEval~\cite{cho2023visual} on multiple dimensions.}
\end{table*}
\begin{figure}[t]
    \centering
    \includegraphics[width=0.45\textwidth]{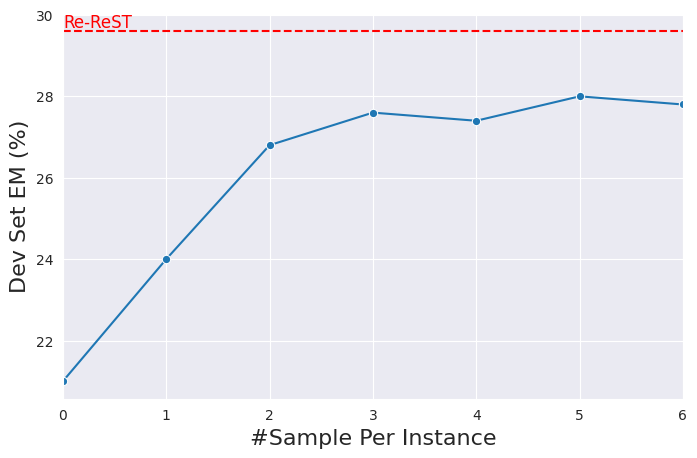}
     \includegraphics[width=0.45\textwidth]{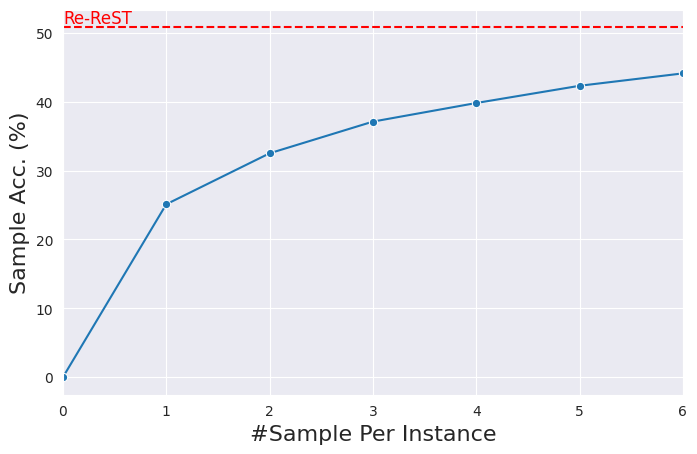}
\caption{\label{fig:hotpotqa-sample} In self-training, increasing the number of generations per instance initially improves model performance, but this effect plateaus. Additionally, both model performance and the number of solved training instances are lower than with Re-ReST, indicating our reflector can efficiently and effectively generate high-quality self-training data. }
\end{figure}

\begin{table}[t]
\begin{tabular}{lcccccc}
\toprule
\multicolumn{1}{c}{\bf Model}  &  \multicolumn{1}{c}{\bf  Sample Acc.} &  \multicolumn{1}{c}{\bf EM} \\
\midrule
Self-Training  & 37.1 & 27.6 \\
Re-ReST w/o Ref. Train. & 43.7 & 28.8 \\
Re-ReST  & 50.8 & 29.6\\
\bottomrule
\end{tabular}
\caption{\label{tab:ablate-reflector} While directly using a pretrained LLM as our reflector improves self-training, training the reflector specifically for self-correction further improves the agent performance.}
\end{table}

\begin{table}[t]
\begin{tabularx}{0.5\textwidth}{
    l
    >{\centering\arraybackslash}X
}
\toprule
\multicolumn{1}{c}{\bf Model}  &  \multicolumn{1}{c}{\bf EM } \\
\midrule
Base  & 27.6 \\
S.C. (6 agents) & 30.8 \\
S.C. (3 agents + 3 reflectors)  & 32.0\\
Oracle (3 agents + 3 reflectors) & 36.8 \\
\bottomrule
\end{tabularx}
\caption{\label{tab:self-consistency} Previous work relies on ground-truth feedback for test-time reflection (Oracle). In contrast, we propose to use self-consistency~\citep{wang2023selfconsistency} to enable our reflector to be applied during inference without ground-truth feedback and achieve improvements, demonstrating the potential of applying our method during the test time.}
\end{table}

\begin{table*}[t]
\small
\begin{tabularx}{1.0\textwidth}{
    l
    >{\centering\arraybackslash}X
    >{\centering\arraybackslash}X
    >{\centering\arraybackslash}X
    >{\centering\arraybackslash}X
    >{\centering\arraybackslash}X
}
\toprule
\multicolumn{1}{c}{\multirow{2}{*}{\bf Model}} &  \multicolumn{1}{c}{\multirow{2}{*}{\bf HotpotQA EM}} &  \multicolumn{1}{c}{\multirow{2}{*}{\bf MBPP Pass@1 }} &  \multicolumn{3}{c}{\bf VPEval Score} \\
& & & \bf Count & \bf Spatial & \bf Scale  \\
\midrule
Self-Training & 27.6 & 54.5  &  74.7  &  54.5 & 29.3   \\
Re-ReST  & 29.6 &  56.4   & 75.0 & 58.2 & 30.1  \\
\midrule
Self-Training w/ DPO & 28.0 & 54.9 & 74.6 & 56.7 & 30.0\\
Re-ReST w/ DPO & 31.0 & 56.4 & 75.4 &  58.5 & 31.0\\
\bottomrule
\end{tabularx}
\caption{\label{results:dpo} Our method is compatible with direct preference optimization (DPO)~\citep{rafailov2024direct}, and integrating DPO into our method can generally improve the model performance.}
\end{table*}
\paragraph{Dataset.} We also conduct experiments in text-to-image generation. Specifically, we use the dataset constructed by~\citet{cho2023visual}. Their dataset evaluates the model's generated images in multiple dimensions and has training data for the spatial, scale, and count dimensions. For each dimension, the evaluation set consists of 1,000 instances. The training dataset consists of 36,920/18,200/1,560 instances for the spatial/scale/count dimensions. 

\paragraph{Model Setup.} We use VPGen in~\citet{cho2023visual} as our base model, which is based on Vicuna-13B~\citep{vicuna2023} and is finetuned for text-to-layout generation on multiple constructed image-text datasets. The generated layouts are fed into an external model (i.e., GLIGEN~\citep{li2023gligen}) for image generation. We build both the agent and reflector upon the VPGen model.

\paragraph{Training and Evaluation Setup.} We use VPGen to perform inference on their training data, and evaluate the generations using VPEval~\citep{cho2023visual}. Specifically, during evaluation, a visual question answering model (BLIP-2~\citep{li2023blip}) is used to determine if the generated images correctly capture the input text information. The BLIP-2 generated results are treated as the environmental feedback for the reflector. We do not use zero-shot reflection results to train the reflector because LLMs cannot perform this task without finetuning. The agent and reflector are trained for 2 epochs with a learning rate of 1e-5 using LoRA. 

\paragraph{Results.} As shown in Table~\ref{results:t2i}, our method continues showing improvements over baselines in the text-to-image generation task. The baseline VPGen model's performance is enhanced when self-training is applied, further improved significantly with our Re-ReST method across all the dimensions. The results demonstrate promising applications of our model in the multimodal generation domain with a language agent as a backend.

\subsection{Analysis}

\paragraph{Re-ReST v.s. Self-Training with More Samples.} We investigate if we can simply sample more generations from the language agent for self-training and achieve comparable performance with our reflector-augmented method. Specifically, we try to sample $k$ generations for each instance, where $k$ is set to $1, 2, 3, 4, 5, 6$, and use the generated samples for self-training. As shown in Figure~\ref{fig:hotpotqa-sample}, if we keep sampling more generations from the language agent, the agent can indeed solve more instances and we can obtain an increasing amount of data for self-training. However, 1) the number of solved instances is still lower than the number of reflector-solved instances, demonstrating that the reflector can find the correct solutions more efficiently than sampling; 2) the model performance is not always improved with more training data and it cannot outperform our method even when trained with more generated samples, indicating that the quality of the self-training data is also important and our reflector can generate training data effectively for the agent.

\paragraph{Effect of Training the Reflector.} As illustrated, we propose to first train the reflector before using it to generate the self-training data. In this part, we investigate if we can use the reflector to perform self-correction in a zero-shot manner and then train the language agent. As in Table~\ref{tab:ablate-reflector}, we find that while the reflector can perform self-correction without any finetuning and improve the performance of the language agent, further improvements can be made if we specifically train the model for self-correction, demonstrating the effectiveness of our proposed reflector training strategy.

\paragraph{Test-Time Reflection without Ground-Truth Feedback.} Previously, our reflector functions only during training and is not used during inference because it is often impossible to obtain ground-truth feedback, which is required for reflection methods to work~\cite{huang2024large}. In this section, we propose employing self-consistency~\citep{wang2023selfconsistency} to enable test-time reflection and address this limitation. Self-consistency is a decoding technique that combines multiple model predictions by sampling various reasoning paths and then selecting the most consistent answer through a majority vote. This approach allows us to apply the reflector during inference. Specifically, we sample multiple answers from our model and perform reflection on each output, regardless of correctness. We then aggregate all the answers using self-consistency. As in Table~\ref{tab:self-consistency}, integrating our reflector with self-consistency (3 agent samples and 3 reflection samples) achieves improvements over baseline (self-consistency with 6 model samples). This demonstrates the potential application of our method during inference, overcoming the current limitation of requiring ground-truth feedback for reflection methods.

\paragraph{Re-ReST with Direct Preference Optimization.} Our reflector turns incorrect samples into correct ones, naturally making negative-positive pairs suitable for preference optimization objectives such as DPO. In this part, we investigate the application of DPO in our method. As in Table~\ref{results:dpo}, integrating DPO into our method can generally improve or achieve comparable performance with training models only with supervised training on positive samples, indicating our compatibility with DPO.

\section{Related Work}


In this section, we first overview the research progress in language agents, then briefly describe self-training and self-correction methods for improving language agents. We also summarize the major differences between our work and previous language agent methods in Table~\ref{tab:work}.

\paragraph{Language Agents.} Language agents refer to language models that interact with the world in general. It has been demonstrated that LLMs can perform actions by generating specific commands~\citep{nakano2021webgpt,huang2022language,ahn2022can} and calling external tool APIs~\citep{lu2024chameleon,schick2024toolformer,gou2024tora}. By integrating the model reasoning and acting abilities, ReAct~\citep{yao2022react} asks an LLM to first generate reasoning traces and then act accordingly, which is then improved by follow-up works through inference-time techniques such as reflection~\citep{shinn2023reflexion} and planning~\citep{yao2024tree, yang2023lacma}. Recently, finetuning agents~\citep{chen2023fireact,yin2024lumos} have attracted attention from the research community. However, most of the existing works attempt to distill knowledge from a relatively strong LLM (e.g., GPT-4) to a weaker LLM (e.g., LLaMa-2). By contrast, our work bootstraps a language agent's performance by utilizing its own reflective ability without using external models.

\paragraph{Self-Training for Language Models.} Various self-training algorithms have been proposed to improve language models~\citep{he2019revisiting,huang2023large,dong2023raft,gulcehre2023reinforced,yuan2024self}, with the general idea being to improve models with self-generated samples in an unsupervised or semi-supervised manner.~\citet{he2019revisiting} is one early work in applying self-training to generative language models and points out the importance of introducing noises during pseudo-label generation to increase the sample diversity. In the large language model era, ~\citet{gulcehre2023reinforced} propose Reinforced Self-Training (ReST), where they use a scoring function to select self-generated samples and augment the training data. Similarly,~\citet{yuan2024self} proposes self-rewarding that scores samples with the LLM itself and trains the model with direct preference optimization (DPO)~\citep{rafailov2024direct} on the scored samples. Self-training has also been employed to improve the chain-of-thought reasoning~\citep{nye2022show,wei2022chain} ability of LLMs~\citep{uesato2022solving}. For example, ~\citet{zelikman2022star} propose to ask an LLM to generate rationales given questions and improve the LLM with its own generated reasoning. Re-ReST falls under the self-training paradigm, and different from previous work, our aim is to generate useful samples efficiently for self-training.

\paragraph{Self-Reflection/Self-Correction for Language Models.} Several works have used LLMs to reflect on their generations with internal or external feedback and correct their errors~\citep{welleck2023generating,wang2023enable,shinn2023reflexion,madaan2023self,kim2024language,ji2024aligner}. A majority of this line of research is focused on improving LLMs during inference. For example, Self-Refine~\citep{madaan2023self} proposes to have LLMs iteratively evaluate their generations, based on which they improve their generations. Similarly,~\citet{shinn2023reflexion} use LLM agents to reflect on its generations and their environment feedback, then guide the next generation with the generated verbal feedback. As pointed out by~\citet{huang2024large},  high-quality external feedback is essential for these self-correction models, without which existing techniques actually decrease model performance. However, such high-quality feedback is often unavailable during the test time, thus we propose to use Re-ReST only during training and perform corrections with oracle feedback from environments, ensuring its effectiveness in correcting the model generations. In addition, during the test time, the corrected generations are distilled into the language model, thus directly generating the answer without introducing overhead during inference.

\section{Conclusion}
Our study studies the applications of self-training in language agents and improves it with Reflection-Reinforced Self-Training (Re-ReST), an approach that efficiently obtains high-quality samples for self-training with a reflector. Our experiments demonstrate that Re-ReST outperforms self-training methods across various tasks, confirming the efficiency and effectiveness of incorporating a reflection mechanism. Within the proposed framework, in the future, we can improve the reflection mechanism and develop better training paradigms for the agent and reflector. 

\section*{Acknowledgement}
We thank anonymous reviewers and UCLA NLP group members for their insightful feedback. This research is based upon work supported by DARPA ECOLE Program No. \#HR00112390060, OFFICE OF NAVAL RESEARCH Award \#N00014-23-1-2780, an Amazon AGI foundation research award, a google research scholar grant, and CISCO sponsored research award.

\section*{Limitations}
Our approach is predicated on the availability of ground-truth feedback during the training process. While this assumption holds true for many language agent tasks, it presents challenges when applied to broader contexts. Specifically, acquiring accurate ground-truth feedback can be difficult in diverse, real-world scenarios. This limitation underscores a key aspect of our study: it is primarily concentrated on language agent tasks, thereby neglecting the potential applications and implications within the broader scope of general language modeling. This suggests the need for future research to explore and address the complexities of applying our methods to general language modeling tasks, where ground-truth feedback may not be as readily accessible or reliable. Another potential risk of the method is that through self-training, the biases encoded in LLMs can be amplified, and careful calibrations should be conducted before the deployment of our method.

\bibliography{custom}

\clearpage

\appendix


\begin{table*}[h!]
\begin{center}
\setlength{\tabcolsep}{3.0pt}
\begin{tabular}{lcccccccc}
\toprule
\multicolumn{1}{c}{\multirow{2}{*}{\bf Work}}  &  \multicolumn{2}{c}{\multirow{1}{*}{\bf Agent Training}}  &  \multicolumn{2}{c}{\bf  Agent Reflection} \\
& \bf Finetuning  & \bf GPT-Free &  \bf w/ G.T. Feedback  & \bf w/o G.T. Feedback \\
\midrule
FireAct~\cite{chen2023fireact} & \cmark  & \xmark & \xmark  & \xmark \\
LUMOS~\cite{yin2024lumos} & \cmark  & \xmark & \xmark  & \xmark   \\
Reflexion~\cite{shinn2023reflexion} & \xmark  & \xmark & \cmark  & \xmark    \\
Self-Refine~\cite{madaan2023self} & \xmark  & \xmark & \cmark  & \xmark    \\
Re-ReST & \cmark    & \cmark & \cmark & \cmark  \\
\bottomrule
\end{tabular}
\end{center}
\caption{\label{tab:work} Comparisons with previous language agent methods. We propose to finetune LLMs for language agent tasks with self-generated data, while previous work such as FireAct and LUMOS rely on stronger LLMs such as GPT-4 to perform knowledge distillation. In addition, we propose to use the agent's reflection ability to improve the self-training efficiency, where the reflection can function both with and without ground-truth feedback, addressing the limitation of previous agent reflection methods~\cite{shinn2023reflexion,madaan2023self,huang2024large}.}
\end{table*}



\begin{table*}[h!]\centering
\begin{minipage}{0.95\textwidth}  
\centering
\begin{tcolorbox} 
    \centering
    \small
    \begin{tabular}{p{0.95\textwidth}}
   { {\bf Prompt} } \\
Solve a question answering task with interleaving Thought, Action, Observation steps. Thought can reason about the current situation, and Action can be three types:\\(1) Search[entity], which searches the exact entity on Wikipedia and returns the first paragraph if it exists. If not, it will return some similar entities to search. \\(2) Lookup[keyword], which returns the next sentence containing keyword in the last passage successfully found by Search.\\(3) Finish[answer], which returns the answer and finishes the task.\\You may take as many steps as necessary.\\
    \midrule
    \textcolor[rgb]{0,0.7,0}{ \{In-context examples\} } \\
    Question: What is the elevation range for the area that the eastern sector of the Colorado orogeny extends into? \\
    Thought 1: I need to search Colorado orogeny, find the area that the eastern sector of the Colorado orogeny extends into, then find the elevation range of the area.\\Action 1: Search[Colorado orogeny]\\Observation 1: The Colorado orogeny was an episode of mountain building (an orogeny) in Colorado and surrounding areas.\\Thought 2: It does not mention the eastern sector. So I need to look up eastern sector.\\Action 2: Lookup[eastern sector]\\Observation 2: (Result 1 / 1) The eastern sector extends into the High Plains and is called the Central Plains orogeny.\\Thought 3: The eastern sector of Colorado orogeny extends into the High Plains. So I need to search High Plains and find its elevation range.\\Action 3: Search[High Plains]\\Observation 3: High Plains refers to one of two distinct land regions:\\Thought 4: I need to instead search High Plains (United States).\\Action 4: Search[High Plains (United States)]\\Observation 4: The High Plains are a subregion of the Great Plains. From east to west, the High Plains rise in elevation from around 1,800 to 7,000 ft (550 to 2,130 m).[3]\\Thought 5: High Plains rise in elevation from around 1,800 to 7,000 ft, so the answer is 1,800 to 7,000 ft.\\Action 5: Finish[1,800 to 7,000 ft] \\
    ... \\
    \midrule
 {\bf User}  \\
    Question: \textcolor[rgb]{0.8,0,0}{\{Input question\}} \\
    \end{tabular}
\end{tcolorbox}
\caption{Prompt template for the HotpotQA agent. \textcolor[rgb]{0,0.7,0}{ \{In-context examples\} }  \textcolor[rgb]{0.8,0,0}{\{Input question\}}  }
\end{minipage}
\end{table*}

\begin{table*}[h!]\centering
\begin{minipage}{0.95\textwidth}  
\centering
\begin{tcolorbox} 
    \centering
    \small
    \begin{tabular}{p{0.95\textwidth}}
   { {\bf Prompt} } \\
Solve a question answering task with a reflection on a previous trial and interleaving Thought, Action, Observation steps. Reflection can reason on a previous trial that failed and develop a new strategy to avoid repeating the same mistake. Thought can reason about the current situation, and Action can be three types: \\(1) Search[entity], which searches the exact entity on Wikipedia and returns the first paragraph if it exists. If not, it will return some similar entities to search.\\(2) Lookup[keyword], which returns the next sentence containing keyword in the last passage successfully found by Search.\\(3) Finish[answer], which returns the answer and finishes the task.\\You need to reflect on the previous trial and then restart a new trial.\\
    \midrule
    \textcolor[rgb]{0,0.7,0}{ \{In-context examples\} } \\
    Question: When was the subject of the book The Deliberate Stranger executed?\\Previous Trial:\\Thought 1: I need to search The Deliberate Stranger and find when the subject was executed.\\Action 1: Search[The Deliberate Stranger]\\Observation 1: The Deliberate Stranger is a book about American serial killer Ted Bundy written by Seattle Times reporter Richard W. Larsen that was published in 1980. The book spawned a television miniseries of the same title, starring Mark Harmon as Bundy, that aired on NBC on May 4, 1986.\\Thought 2: The subject of the book, Ted Bundy, was executed in 1989.\\Action 2: Finish[1989]
    Reflection: I should have searched for Ted Bundy first before I finished answering the question.\\Thought 1: I need to first search The Deliberate Stranger and find its subject, and then find out which year the subject was executed.\\Action 1: Search[The Deliberate Stranger]\\Observation 1: The Deliberate Stranger is a book about American serial killer Ted Bundy written by Seattle Times reporter Richard W. Larsen that was published in 1980. The book spawned a television miniseries of the same title, starring Mark Harmon as Bundy, that aired on NBC on May 4, 1986.\\Thought 2: The subject of The Deliberate Stranger is Ted Bundy. I need to search for Ted Bundy and find out which year he was executed.\\Action 2: Search[Ted Bundy]\\Observation 2: Could not find [Ted Bundy]. Similar: ['Ted Bundy', 'Ted Bundy (film)', 'Ted Bundy: American Boogeyman', 'Conversations with a Killer: The Ted Bundy Tapes', 'Murder of Rita Curran', 'Ted Bundy: Falling for a Killer', 'Extremely Wicked, Shockingly Evil and Vile', 'Disappearance of Ann Marie Burr', 'Murder of Georgann Hawkins', 'Zac Efron']\\Thought 3: I need to change the keyword for search.\\Action 3: Search[Ted Bundy executed]\\Observation 3: Theodore Robert Bundy (Cowell; November 24, 1946 January 24, 1989) was an American serial killer who kidnapped, raped and murdered dozens of young women and girls during the 1970s and possibly earlier. After more than a decade of denials, he confessed to 30 murders committed in seven states between 1974 and 1978. His true victim total is unknown.Bundy often employed charm to disguise his murderous intent when kidnapping victims, and extended this tactic vis-vis law enforcement, the media and the criminal justice system to maintain his claims of innocence... \\ 
    Thought 4: Ted Bundy was executed on January 24, 1989.\\Action 4: Finish[January 24, 1989] \\
    ... \\
    \midrule
 {\bf User}  \\
    Question: \textcolor[rgb]{0.8,0,0}{\{Input question and previous trial\}} \\
    \end{tabular}
\end{tcolorbox}
\caption{Prompt template for the HotpotQA reflector. \textcolor[rgb]{0,0.7,0}{ \{In-context examples\} }  \textcolor[rgb]{0.8,0,0}{\{Input question and previous trial\}}  }
\end{minipage}
\end{table*}

\begin{table*}[h!]\centering
\begin{minipage}{0.95\textwidth}  
\centering
\begin{tcolorbox} 
    \centering
    \small
    \begin{tabular}{p{0.95\textwidth}}
   { {\bf Prompt} } \\
You are an expert Python programmer, and here is your task: \textcolor[rgb]{0.8,0,0}{\{input task\}}\\Your code should pass these tests:\\ \textcolor[rgb]{0,0.7,0}{ \{unit tests\} } \\Your code should start with a [PYTHON] tag and end with a [/PYTHON] tag.
    \end{tabular}
\end{tcolorbox}
\caption{Prompt template for the MBPP agent. \textcolor[rgb]{0,0.7,0}{ \{unit tests\} }  \textcolor[rgb]{0.8,0,0}{\{input task\}}  }
\end{minipage}
\end{table*}

\begin{table*}[h!]\centering
\begin{minipage}{0.95\textwidth}  
\centering
\begin{tcolorbox} 
    \centering
    \small
    \begin{tabular}{p{0.95\textwidth}}
   { {\bf Prompt} } \\
You are an AI Python assistant. You will be given the user input, your past incorrect function implementation, and a series of unit tests. Write your reflection on the function implementation and correct your implementation (copy the function signature and its docstring). \\
\midrule
 \textcolor[rgb]{0,0.7,0}{ \{In-context examples\} } \\
 
[previous impl]: \\
```python \\
def add(a: int, b: int): \\
    """ \\
    Given integers a and b, return the total value of a and b. \\
    """ \\
    return a - b \\
``` \\

[unit test results from previous impl]: \\
Tested passed: \\

Tests failed: \\
assert add(1, 2) == 3 \# output: -1 \\ 
assert add(1, 2) == 4 \# output: -1 \\

[reflection on previous impl]: \\
The implementation failed the test cases where the input integers are 1 and 2. The issue arises because the code does not add the two integers together, but instead subtracts the second integer from the first. To fix this issue, we should change the operator from `-` to `+` in the return statement. This will ensure that the function returns the correct output for the given input. \\

[improved impl]: \\
```python \\
def add(a: int, b: int): \\
    """ \\
    Given integers a and b, return the total value of a and b. \\
    """ \\
    return a + b \\
```''' \\
 ... \\
 \midrule
 {\bf User}  \\
    \textcolor[rgb]{0.8,0,0}{\{Input task and previous trial\}} \\
 \midrule
    \end{tabular}
\end{tcolorbox}
\caption{Prompt template for the MBPP reflector. \textcolor[rgb]{0,0.7,0}{ \{In-context examples\} }  \textcolor[rgb]{0.8,0,0}{\{Input task and previous trial\}}  }
\end{minipage}
\end{table*}

\begin{table*}[h!]\centering
\begin{minipage}{0.95\textwidth}  
\centering
\begin{tcolorbox} 
    \centering
    \small
    \begin{tabular}{p{0.95\textwidth}}
   { {\bf Prompt} } \\
class ImagePatch: \\
~~~~"""A Python class containing a crop of an image centered around a particular object, as well as relevant information. \\
~~~~Methods \\
~~~~------- \\
~~~~find(object\_name: str)-$>$List[ImagePatch] \\
~~~~~~~~Returns a list of new ImagePatch objects containing crops of the image centered around any objects found in the image matching the object\_name. \\
~~~~simple\_query(question: str=None)-$>$str \\
~~~~~~~~Returns the answer to a basic question asked about the image. If no question is provided, returns the answer to "What is this?". \\
~~~~exists(object\_name: str)-$>$bool \\
~~~~~~~~Returns True if the object specified by object\_name is found in the image, and False otherwise. \\
~~~~verify\_property(property: str)-$>$bool \\
~~~~~~~~Returns True if the property is met, and False otherwise. \\
~~~~best\_text\_match(string1: str, string2: str)-$>$str \\
~~~~~~~~Returns the string that best matches the image.
~~~~crop(left: int, lower: int, right: int, upper: int)-$>$ImagePatch \\
~~~~~~~~Returns a new ImagePatch object containing a crop of the image at the given coordinates. \\
~~~~""" \\
~~~~\textcolor[rgb]{0,0,0.8}{ \{Detailed API definition\} } \\
 \\
\textcolor[rgb]{0,0.7,0}{ \{In-context examples\} } \\
 \\
\textcolor[rgb]{0.8,0,0}{\{Input question\}}
    \end{tabular}
\end{tcolorbox}
\caption{Prompt template for the GQA agent. Full prompt is released in \url{https://github.com/cvlab-columbia/viper/blob/main/prompts/benchmarks/gqa.prompt}. \textcolor[rgb]{0,0,0.8}{ \{Detailed API definition\} } \textcolor[rgb]{0,0.7,0}{ \{In-context examples\} }  \textcolor[rgb]{0.8,0,0}{\{Input question\}}  }
\end{minipage}
\end{table*}

\begin{table*}[h!]\centering
\begin{minipage}{0.95\textwidth}  
\centering
\begin{tcolorbox} 
    \centering
    \small
    \begin{tabular}{p{0.95\textwidth}}
   { {\bf Prompt} } \\
I am writing code to handle visual question answering tasks by calling computer vision APIs. My code is wrong, and I hope you can help correct it. \\
\\
\textcolor[rgb]{0.8,0,0}{\{Input question and previous trial\}} \\
\\
Your response should start with your reasoning and analysis. Then, you should write the correct code wrapped in \`{}\`{}\`{} python and \`{}\`{}\`{}. The correct code should be a function with signature \`{}def execute\_command(image) -$>$ str:\`{} \\
\\
---\\
\\
Below are the available APIs and some example usages:\\
\`{}\`{}\`{}python\\
class ImagePatch: \\
~~~~"""A Python class containing a crop of an image centered around a particular object, as well as relevant information. \\
~~~~Methods \\
~~~~------- \\
~~~~find(object\_name: str)-$>$List[ImagePatch] \\
~~~~~~~~Returns a list of new ImagePatch objects containing crops of the image centered around any objects found in the image matching the object\_name. \\
~~~~simple\_query(question: str=None)-$>$str \\
~~~~~~~~Returns the answer to a basic question asked about the image. If no question is provided, returns the answer to "What is this?". \\
~~~~exists(object\_name: str)-$>$bool \\
~~~~~~~~Returns True if the object specified by object\_name is found in the image, and False otherwise. \\
~~~~verify\_property(property: str)-$>$bool \\
~~~~~~~~Returns True if the property is met, and False otherwise. \\
~~~~best\_text\_match(string1: str, string2: str)-$>$str \\
~~~~~~~~Returns the string that best matches the image.
~~~~crop(left: int, lower: int, right: int, upper: int)-$>$ImagePatch \\
~~~~~~~~Returns a new ImagePatch object containing a crop of the image at the given coordinates. \\
~~~~""" \\
~~~~\textcolor[rgb]{0,0,0.8}{ \{Detailed API definition\} } \\
 \\
\textcolor[rgb]{0,0.7,0}{ \{In-context examples\} } \\
\`{}\`{}\`{}
    \end{tabular}
\end{tcolorbox}
\caption{Prompt template for the GQA reflector. \textcolor[rgb]{0,0,0.8}{ \{Detailed API definition\} }
\textcolor[rgb]{0,0.7,0}{ \{In-context examples\} }  \textcolor[rgb]{0.8,0,0}{\{Input question and previous trial\}}  }
\end{minipage}
\end{table*}

\begin{table*}[h!]\centering
\begin{minipage}{0.95\textwidth}  
\centering
\begin{tcolorbox} 
    \centering
    \small
    \begin{tabular}{p{0.95\textwidth}}
   { {\bf Prompt} } \\
Interact with a household to solve a task. Here is an example. \\
    \midrule
    \textcolor[rgb]{0,0.7,0}{ \{In-context examples\} } \\
    You are in the middle of a room. Looking quickly around you, you see a cabinet 4, a cabinet 3, a cabinet 2, a cabinet 1, a countertop 1, a garbagecan 1, a handtowelholder 2, a handtowelholder 1, a sinkbasin 2, a sinkbasin 1, a toilet 1, a toiletpaperhanger 1, and a towelholder 1. \\
    Your task is to: put some spraybottle on toilet. \\
    $>$ think: To solve the task, I need to find and take a sparybottle, then put it on toilet. \\
    OK. \\
    $>$ think: First I need to find a spraybottle. A spraybottle is more likely to appear in cabinet (1-4), countertop (1), toilet (1), sinkbasin (1-2), garbagecan (1). I can check one by one, starting with cabinet 1. \\
    OK. \\
    $>$ go to cabinet 1 \\
    On the cabinet 1, you see a cloth 1, a soapbar 1, a soapbottle 1. \\
    $>$ go to cabinet 2 \\
    The cabinet 2 is closed. \\ 
    $>$ open cabinet 2 \\
    You open the cabinet 2. The cabinet 2 is open. In it, you see a candle 1, and a spraybottle 2. \\
    $>$ think: Now I find a spraybottle (2). Next, I need to take it. \\
    OK \\
    $>$ take spraybottle 2 from cabinet 2 \\
    You pick up the spraybottle 2 from the cabinet 2. \\
    $>$ think: Now I take a spraybottle (2). Next, I need to put it in/on toilet 1. \\
    OK. \\
    $>$ go to toilet 1 \\
    On the toilet 1, you see a soapbottle 2. \\
    $>$ put spraybottle 2 in/on toilet 1 \\
    You put the spraybottle 2 in/on the toilet 1. \\
    \midrule
 {\bf User}  \\
    \textcolor[rgb]{0.8,0,0}{\{Input question\}} Here is the task:
    You are in the middle of a room. Looking quickly around you, you see a bed 1, a desk 2, a desk 1, a drawer 6, a drawer 5, a drawer 4, a drawer 3, a drawer 2, a drawer 1, a garbagecan 1, a laundryhamper 1, a safe 1, a shelf 6, a shelf 5, a shelf 4, a shelf 3, a shelf 2, and a shelf 1. \\
    Your task is to: examine the bowl with the desklamp. \\
    \textcolor[rgb]{0,0,0.8}{\{Reflection Results\}} think: I was stuck in a loop in which I continually picked up the alarmclock 1 instead of turning on the desklamp.    
\\
    \end{tabular}
\end{tcolorbox}
\caption{Example Prompt Template on the ALFWorld dataset. A prompt includes (a) \textcolor[rgb]{0,0.7,0}{ \{In-context example\} } which is a complete trajectory from a successful trial. (b) \textcolor[rgb]{0.8,0,0}{\{Input question\}} describes the initial environment and the instruction of the task, and (c) \textcolor[rgb]{0,0,0.8}{\{Reflection Results\}} encapsulates the self-reflection results from the reflector model.}
\end{minipage}
\end{table*}

\end{document}